\documentclass[10pt, twocolumn, journal,web]{article}
\usepackage{times}
\usepackage{amsmath,amssymb,amsfonts}
\usepackage{subfig}
\usepackage[utf8]{inputenc}
\usepackage{utfsym}
\usepackage{multirow}
\usepackage[normalem]{ulem}
\usepackage{amsmath,amssymb,amsfonts}
\usepackage{algorithmic}
\useunder{\uline}{\ul}{}
\usepackage{url}
\def\BibTeX{{\rm B\kern-.05em{\sc i\kern-.025em b}\kern-.08em
    T\kern-.1667em\lower.7ex\hbox{E}\kern-.125emX}}
\usepackage{graphicx}
\usepackage{authblk}

\usepackage{xcolor}
\usepackage{hyperref}
\usepackage{booktabs}
\hypersetup{
   colorlinks=true,
    linkcolor=blue,
   citecolor=blue,
    urlcolor=blue
}
\usepackage{makecell}
\usepackage{tabularx}

\makeatletter
\renewcommand\AB@affilsepx{, \protect\Affilfont}
\makeatother
\usepackage{tabularx}
\providecommand{\keywords}[1]
{
  \small	
  \textbf{\textit{Keywords---}} #1
}

\begin{document}

\title{\textbf{Comprehensive Benchmarking of YOLOv11 Architectures for Scalable and Granular Peripheral Blood Cell Detection}}
\author[1, 3, 4]{Mohamad Abou Ali}
\author[3, 4]{Mariam Abdulfattah}
\author[3, 4]{Baraah Al Hussein}
\author[1, 2]{Fadi Dornaika\thanks{Corresponding author}}
\author[3, 4]{Ali Cherry}
\author[3, 4]{Mohamad Hajj-Hassan}
\author[3, 4]{Lara Hamawy}

\affil[1]{\textit{University of the Basque Country}}
\affil[2]{\textit{IKERBASQUE}}
\affil[3]{\textit{Lebanese International University (LIU)}}
\affil[4]{\textit{The International University of Beirut}}

\affil[ ]{

\small\texttt{mohamad.abouali01@liu.edu.lb, 22030274@students.liu.edu.lb, 21930849@students.liu.edu.lb, fadi.dornaika@ehu.eus, ali.cherry@liu.edu.lb, mohamad.hajjhassan@liu.edu.lb, lara.hamawy@liu.edu.lb}}
\date{}
\maketitle
\begin{abstract}
Manual peripheral blood smear (PBS) analysis is labor-intensive and subjective. While deep learning offers a promising alternative, a systematic evaluation of state-of-the-art models such as YOLOv11 for fine-grained PBS detection is still lacking. In this work, we make two key contributions. First, we curate a large-scale annotated dataset for blood cell detection and classification, comprising 16,891 images across 12 peripheral blood cell (PBC) classes, along with the red blood cell class, all carefully re-annotated for object detection tasks. In total, the dataset contains 298,850 annotated cells. Second, we leverage this dataset to conduct a comprehensive evaluation of five YOLOv11 variants (ranging from Nano to XLarge). These models are rigorously benchmarked under two data-splitting strategies (70:20:10 and 80:10:10) and systematically assessed using multiple performance criteria, including mean Average Precision (mAP), precision, recall, F1-score, and computational efficiency. Our experiments show that the YOLOv11-Medium variant achieves the best trade-off, reaching a mAP@0.5 of 0.934 under the 8:1:1 split. Larger models (Large and XLarge) provide only marginal accuracy gains at substantially higher computational cost. Moreover, the 8:1:1 split consistently outperforms the 7:2:1 split across all models.
These findings highlight YOLOv11—particularly the Medium variant—as a highly effective framework for automated, fine-grained PBS detection. Beyond benchmarking, our publicly released dataset (\textcolor{blue}{https://github.com/Mohamad-AbouAli/OI-PBC-Dataset}) offers a valuable resource to advance research on blood cell detection and classification in hematology.
\end{abstract}

 \hspace{10pt}
\section{Introduction}
\label{sec:Intro}

Peripheral blood cell (PBC) analysis plays a pivotal role in hematology, offering critical insights for the diagnosis and monitoring of infections, hematological malignancies, and immune system disorders ~\cite{hoffbrand2023essential}. Traditionally, this process relies on manual microscopic examination of stained blood smears a method that, while clinically effective, is inherently constrained by observer variability, fatigue, and the time-intensive expertise required ~\cite{hegde2019automated}. These limitations underscore the urgent need for automated, reliable systems that can assist clinicians in delivering faster and more consistent diagnostic outcomes.

Recent advances in artificial intelligence (AI) and computer vision, particularly deep learning, have revolutionized medical image analysis. These technologies are increasingly being applied to hematology for tasks including cell detection, segmentation, and classification~\cite{chen2021recent}. However, the development of robust automated blood cell analysis systems faces two primary obstacles: (1) a scarcity of high-quality, publicly available datasets with detailed annotations for the full spectrum of clinically relevant cell types, and (2) the need for a systematic evaluation of modern, efficient object detection architectures tailored for this specific domain. Challenges such as high visual similarity among PBC subtypes, significant class imbalance—especially for rare cell types—and the necessity for computational efficiency in clinical settings further complicate progress~\cite{zhang2021challenges,abouali2023blood}. To address severe class imbalance, we introduced a novel data augmentation technique, Naturalize ~\cite{ali2024advancing}, which was further validated in other medical imaging applications~\cite{abouali2024naturalize}. 

This study directly addresses these gaps and makes two core contributions:

First, building on an existing dataset~\cite{acevedo2020dataset} originally dedicated to PBC classification, we curate a meticulously annotated resource for the detection of twelve distinct peripheral blood cell types. The resulting large-scale dataset comprises 16,891 images encompassing 12 peripheral blood cell (PBC) classes as well as the red blood cell class, all re-annotated with bounding boxes for object detection and classification tasks. This curated dataset offers a valuable benchmark to support and advance future research in automated blood cell analysis.

Second, we conduct a comprehensive and systematic evaluation of the recently released YOLOv11 architecture~\cite{Khanam2024YOLOv11} for the task of blood cell detection. While convolutional neural network (CNN)-based classifiers have demonstrated strong performance, object detection models provide greater versatility by simultaneously localizing and classifying multiple cells within a single image. Frameworks such as Faster R-CNN, YOLO, and RetinaNet~\cite{Lin2018focallossdenseobject, Wei2022} have achieved notable success; however, the adoption of the latest YOLOv11 model—incorporating architectural advancements that improve both speed and accuracy—remains largely unexplored and insufficiently evaluated in hematology~\cite{yurdakul2025bc}. To this end, we conduct a rigorous comparative analysis across five YOLOv11 variants (Nano, Small, Medium, Large, and XLarge) using our novel dataset. Performance is benchmarked using key metrics including mean Average Precision (mAP), precision, recall, F1-score, training duration, and inference speed. Our findings offer practical insights into the trade-offs between detection accuracy and computational efficiency, providing a foundational guide for developing robust and clinically applicable automated PBC analysis systems.

The remainder of this paper is structured as follows. Section \ref{sec:previous} surveys related work in traditional and deep learning-based blood cell analysis. Section \ref{sec:methods} describes our comprehensive methodology, covering dataset curation, class refinement, annotation, and the YOLOv11 model family. Our experimental results and a detailed comparative analysis of all models are presented in Section \ref{sec:results}. In addition, Section \ref{sec:results} discusses the clinical implications of our findings, acknowledges the study's limitations, and suggests directions for future research. Finally, Section \ref{sec:conclusion} provides concluding remarks.

\section{Previous Works}
\label{sec:previous}

Automated analysis of Peripheral Blood Smear (PBS) images has evolved significantly, transitioning from traditional machine learning approaches to advanced deep learning and transformer-based architectures. This progression reflects the growing demand for faster, more accurate, and standardized diagnostic tools in hematology, addressing the limitations of manual microscopic examination.
\subsection{Traditional Machine Learning Approaches}

Early efforts in blood cell analysis relied on handcrafted features—morphometric, textural, and colorimetric descriptors—combined with classifiers such as Support Vector Machines (SVMs) and decision trees. Ushizima et al. ~\cite{ushizima2005support} developed an SVM-based system to classify six white blood cell (WBC) types, including malignant chronic lymphocytic leukemia (CLL) cells, achieving 90\% accuracy using a tree-based multiclass strategy. Dsilva et al. ~\cite{dsilva2023wavelet} applied the Wavelet Scattering Transform (WST) with SVMs for dengue diagnosis, reaching 98.7\% accuracy. However, these methods lacked scalability due to their dependence on manual feature engineering. In the same study, YOLOv8 was introduced for object detection, outperforming WST-SVM with a mean accuracy of 99.3\%, underscoring the superiority of deep learning for heterogeneous datasets.
\subsection{Deep Learning for Blood Cell Detection and Classification}
Deep learning has revolutionized PBC analysis by enabling models to learn hierarchical features directly from raw image data. Jiang et al. ~\cite{jiang2021improved} enhanced YOLOv3 with attention mechanisms, improving mAP from 0.872 to 0.943 on the BCCD dataset ~\cite{Lee2022Complete} . Chen et al.~\cite{liu2022accurate}  integrated ResNet and DenseNet with a Spatial and Channel Attention Module (SCAM), achieving 97.84\% accuracy across multiple datasets. Yao et al.~\cite{yao2021high} compared Faster R-CNN and YOLOv4, noting YOLOv4’s real-time performance at 60 FPS. Rehman et al.~\cite{rehman2023largescale} introduced AttriDet, extending YOLOv5 with attribute prediction capabilities. Guo et al.~\cite{zhang2023blood} further optimized YOLOv5 using attention modules and refining its bounding box regression loss function to improve detection accuracy, while Gan et al.~\cite{gan2021txl} proposed the TXL-PBC dataset with semi-automatic annotation via YOLOv8n, achieving mAP@0.5 of 97.0\%.
\subsection{Transformer-Based and Hybrid Architectures}
Recent work has explored transformer-based models for blood cell detection, leveraging self-attention to capture global context. Nugraha et al.~\cite{erfianto2023white} combined YOLOv8 with Detection Transformer (DETR), significantly improving basophil and lymphocyte detection. Li et al.~\cite{li2023peripheral} utilized Fourier Ptychographic Microscopy (FPM) and Deep Convolutional GANs (DCGANs) to enhance dataset quality, achieving an mAP of 0.936 with DETR. Chen et al.~\cite{chen2024accurate} proposed MFDS-DETR, a hybrid CNN–Transformer model with multi-scale deformable attention, reaching AP50 scores up to 99.9\% on dense cell datasets. Table \ref{tab:previous_studies}  provides a comprehensive assessment of recent advancements in automated blood cell detection. In this table,  we present a comparison  summarizing the strengths and limitations of prior studies. This table consolidates key performance indicators—including mAP, accuracy, precision, recall, and F1-score—across various machine learning, deep learning, and transformer-based approaches applied to peripheral blood smear analysis.

All previous works in Table \ref{tab:previous_studies} are on the BCCD dataset, which is limited by the number of images "364 images", the number of classes "3" (WBC, RBC and Platelet), and the image quality.

\begin{table*}[htbp] 
\centering 
\caption{Performance Comparison of Blood Cell Detection Methods} \label{tab:previous_studies} 
\resizebox{\textwidth}{!}{ 
\begin{tabular}{lcccccc} \hline \textbf{Paper} & \textbf{mAP} & \textbf{Accuracy (\%)} & \textbf{Precision (\%)} & \textbf{Recall (\%)} & \textbf{F1 (\%)} \\ \hline Ushizima et al.~\cite{ushizima2005support} & N/A & 94.9 & N/A & N/A & N/A \\ Dsilva et al.~\cite{dsilva2023wavelet} & mAP50-95 of 94.1 & 99.3 ± 1.4 &98.6 & 97.55 & 98.76 \\ Jiang et al.~\cite{jiang2021improved} & 94.3 &RBCs=97.44/WBCs=99.46/ Platelets=96.99 & N/A & N/A & N/A \\ Chen et al.~\cite{chen2021recent} & 93.51 & 97.42 & N/A & 92.35 & 92.66 \\ Yao et al.~\cite{yao2021high} & N/A & 96.25 & High & High & High \\ Rehman et al.~\cite{rehman2023largescale} & 44.2 & N/A & N/A & N/A & 28.2\\ Guo et al~\cite{zhang2023blood} & 97.4 & 99.4 & 97.9 & 93.5 & High \\ Gan et al.~\cite{gan2021txl} & 97.4 & N/A & 96.4 & 94.6 & 95.4 \\ Nughaha~\cite{erfianto2023white} & N/A & N/A & 83.24 & 77.8 & 78.65 \\ Li et al.~\cite{li2023peripheral} & 0.936 for DETR & N/A & N/A & N/A & N/A \\ Chen et al.~\cite{chen2024accurate} &79.7 (WBCDD)/97.2 (AP50)& N/A & N/A & N/A & N/A \\ \hline 
\end{tabular} } 
\end{table*}

While the aforementioned studies demonstrate significant progress in automating PBC analysis, several key limitations persist. Many approaches are evaluated on limited cell taxonomies, often overlooking clinically relevant but morphologically similar sub-classes (e.g., banded vs. segmented neutrophils, or the spectrum of immature granulocytes). Furthermore, the performance of newer, state-of-the-art architectures like the YOLOv11 family remains largely unexplored for this specific domain, particularly in a comprehensive framework that evaluates the trade-off between accuracy and computational efficiency across multiple variants. Most critically, there is a lack of systematic benchmarking that can guide clinical deployment decisions based on real-world constraints. To address these gaps, this paper presents a thorough evaluation of the entire YOLOv11 architecture family (Nano, Small, Medium, Large, and XLarge) for the detection of an expanded 12-class PBC taxonomy. Our novel contribution lies in the meticulous re-annotation of a large-scale public dataset for object detection and a rigorous comparative analysis across two data splitting strategies. This work provides much-needed practical insights into model selection, establishing the YOLOv11-Medium variant as the optimal architecture for achieving a critical balance between high diagnostic accuracy and operational efficiency in clinical settings.


\section{Materials and Methods}
\label{sec:methods}

\begin{figure}[htbp]
\centering
\includegraphics[width=0.95\linewidth,height=0.95\textheight,keepaspectratio]{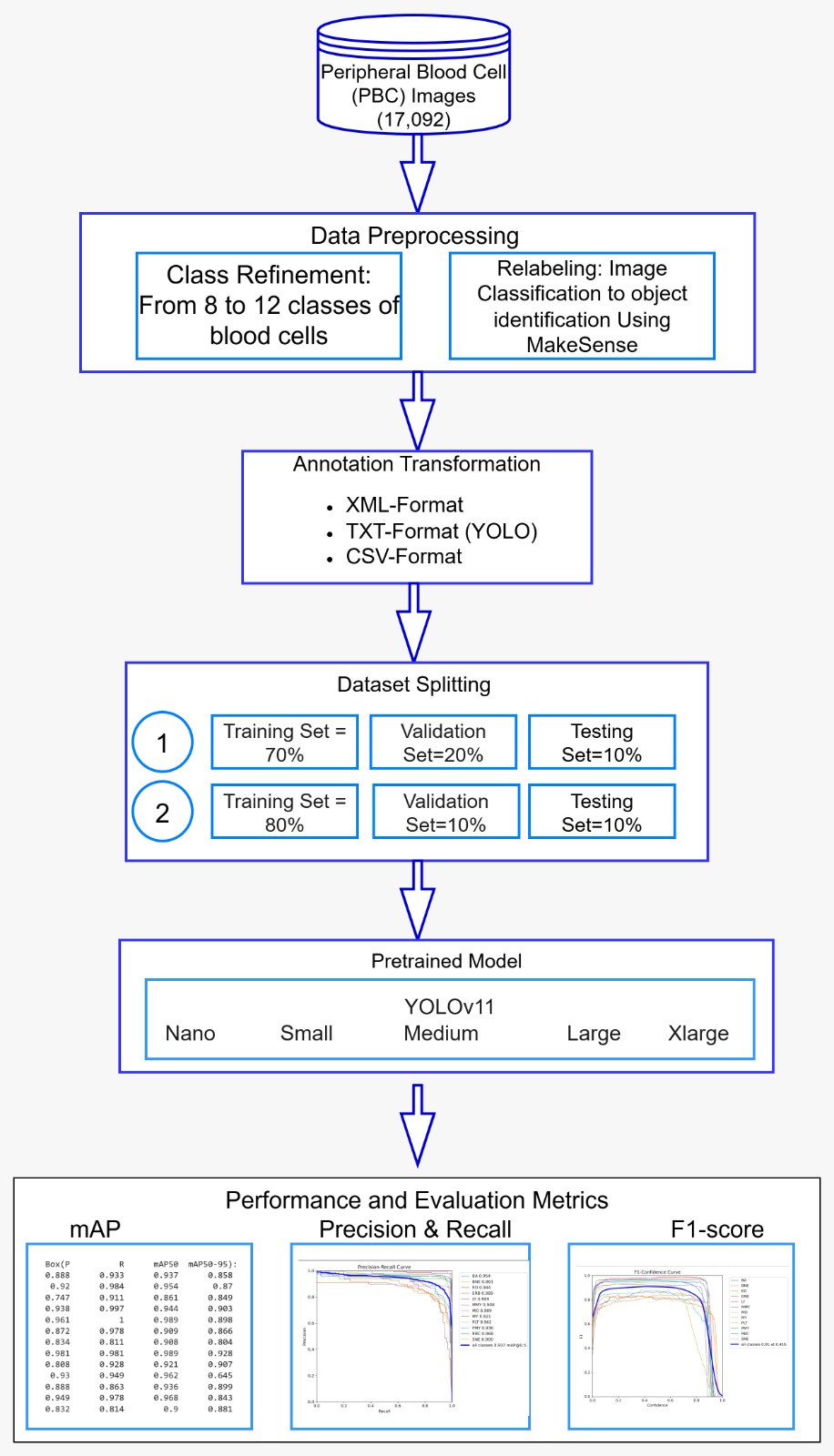}
\caption{Comprehensive pipeline for automated PBC analysis, showing sequential stages from dataset preparation to model evaluation.\label{Figure 2}}
\end{figure}

This study introduces a comprehensive methodological framework for automated peripheral blood smear (PBC) analysis, encompassing critical stages from dataset curation to model evaluation. Our approach is designed for robustness and clinical relevance, beginning with the acquisition and significant refinement of a public microscopic dataset. To enhance diagnostic granularity, we expanded the original 8-class taxonomy into a more clinically meaningful 12-class system. Each cell instance in this refined dataset was meticulously re-annotated at the object level to facilitate precise localization and classification. To rigorously assess model generalization, we employed two distinct dataset splitting strategies (7:2:1 and 8:1:1). Subsequently, a full suite of YOLOv11 model variants—from the lightweight Nano to the high-performance XLarge—were trained and optimized using advanced augmentation techniques. Finally, model performance was benchmarked exhaustively using standard object detection metrics. The complete end-to-end pipeline is illustrated in Figure \ref{Figure 2}. 

\subsection{Peripheral Blood Cell (PBC) Dataset}
\begin{figure}[htbp]
\centering
\includegraphics[width=0.75\linewidth,height=0.75\textheight,keepaspectratio]{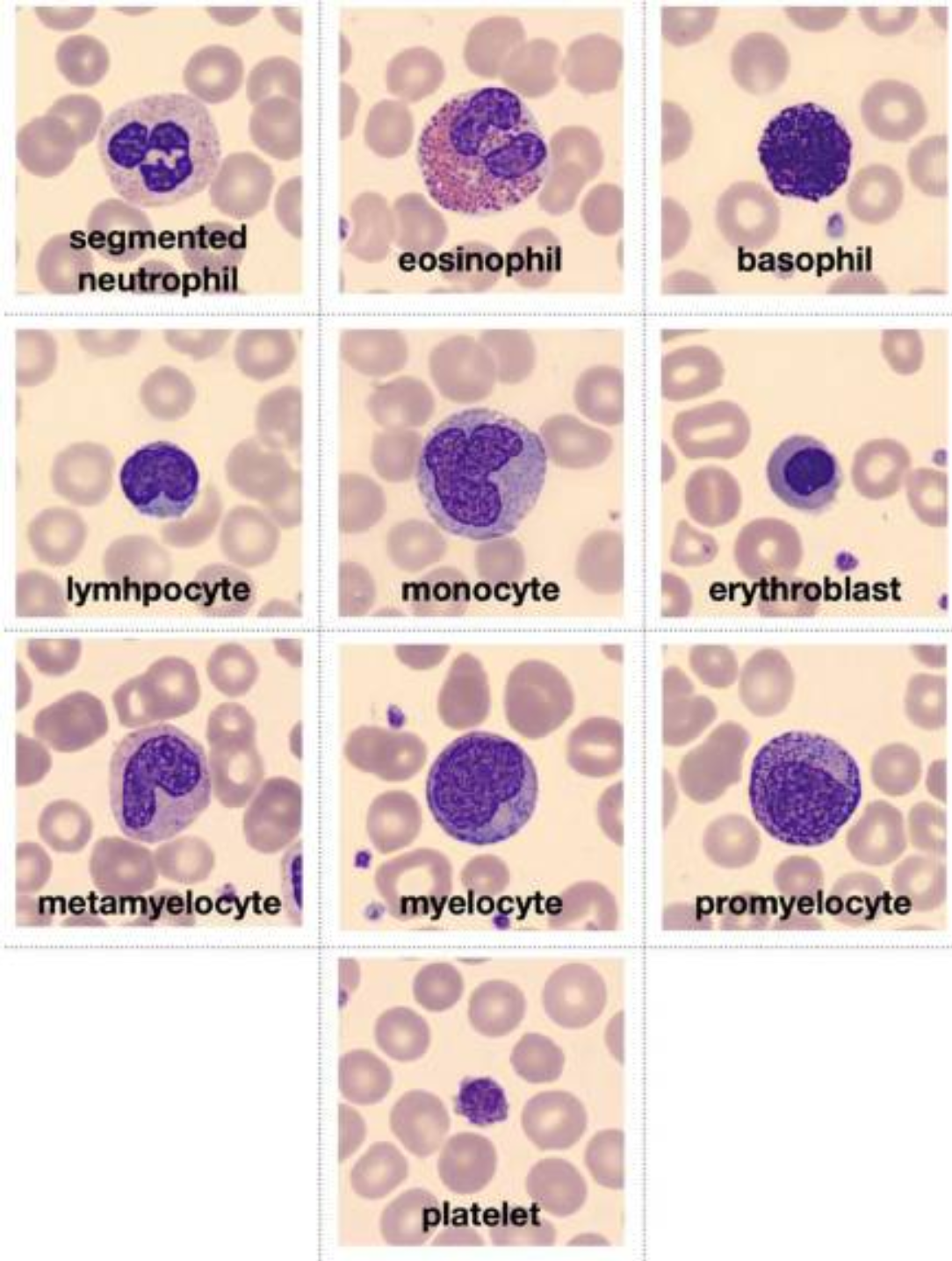}
\caption{Representative examples of the eight original cell categories in the PBC dataset~\cite{acevedo2020dataset}.\textbf{ Note:} the image meta-myelocyte, myelocyte and pro-myelocyte are found under the class Immature Granulocyte in the original PBC dataset.\label{Figure 3}}
\end{figure}

The PBC dataset used in this study is derived from a dataset of microscopic peripheral blood cell images for the development of automatic recognition systems ~\cite{acevedo2020dataset}, acquired using a CellaVision DM96 automated hematology imaging analyzer. Images were captured in RGB color space, stored in JPEG format with a resolution of 360×363 pixels, and annotated by experienced clinical pathologists. The original dataset contains 17,092 images categorized into eight classes: neutrophils, eosinophils, basophils, lymphocytes, monocytes, immature granulocytes (myelocytes, metamyelocytes, and promyelocytes), erythroblasts, and platelets. Representative examples of these classes are shown in Figure \ref{Figure 3}~\cite{acevedo2020dataset}, and their distribution is detailed in Table \ref{tab:cell_distribution}. Each image in the PBC dataset contains one instance belonging to one of the eight classes. This image contains  several red blood cells.

\begin{table*}[htbp]
    \centering
    \caption{Distribution of Cell Types in the Peripheral Blood Cell (PBC) Dataset}
    \label{tab:cell_distribution}
    \begin{tabularx}{\textwidth}{lXrr}
        \toprule
        \textbf{\#} & \textbf{Cell Type} & \textbf{Images} & \textbf{Percent (\%)} \\
        \midrule
        1 & Neutrophils & 3329 & 19.48 \\
        2 & Eosinophils & 3117 & 18.24 \\
        3 & Basophils & 1218 & 7.13 \\
        4 & Lymphocytes & 1214 & 7.10 \\
        5 & Monocytes & 1420 & 8.31 \\
        6 & Immature Granulocytes (IG) & 2895 & 16.94 \\
        7 & Erythroblasts & 1551 & 9.07 \\
        8 & Platelets (Thrombocytes) & 2348 & 13.74 \\
        \midrule
        \multicolumn{2}{l}{\textbf{Total}} & \textbf{17,092} & \textbf{100} \\
        \bottomrule
    \end{tabularx}
\end{table*}

\subsection{Data Preprocessing}
Data preprocessing is a critical phase designed to enhance the quality and utility of the raw PBC images for subsequent machine learning tasks. This stage addresses inherent challenges such as variations in image quality, inconsistencies in cell morphology, and the need for standardized labeling. The preprocessing pipeline involves two primary components: Class Refinement and Relabeling.

\subsubsection{Class Refinement: From 8 to 12 Classes of Blood Cells}
To enhance the clinical relevance and granularity of automated blood cell detection, the dataset was refined from an initial eight classes to a total of twelve distinct categories. This refinement involved two key stages. First, the original eight classes were expanded to eleven distinct blood cell types by sub-dividing neutrophils and immature granulocytes into more specific subcategories : Neutrophils were subdivided into banded neutrophils (BNE) and segmented neutrophils (SNE). Immature granulocytes were further separated into pro-myelocytes (PMY), myelocytes (MY), and meta-myelocytes (MMY). During this refinement process, 151 images from the original dataset were excluded due to insufficient morphological clarity. The refined dataset comprised 16,891 images, with the updated class distribution shown in Table \ref{tab:cell_distribution 11 categories}. As mentioned earlier, each image in the PBC dataset contains one instance belonging to one of the eight classes. This image contains  several red blood cells.

\begin{table*}[htbp]
\centering
\caption{Distribution of Blood Cell Types in the Images After Class Refinement to 11 Categories}
\label{tab:cell_distribution 11 categories}
\begin{tabular}{clccc}
\toprule
\# & \textbf{Cell Type} & \textbf{Symbol} & \textbf{Total Images} & \textbf{Percent (\%)} \\
\midrule
1  & Basophil              & BA  & 1218  & 7  \\
2  & Banded Neutrophil     & BNE & 1633  & 10 \\
3  & Eosinophil            & EO  & 3117  & 18 \\
4  & Erythroblast          & ERB & 1551  & 9  \\
5  & Lymphocyte            & LY  & 1214  & 7  \\
6  & Meta-myelocyte        & MMY & 1015  & 6  \\
7  & Monocyte              & MO  & 1420  & 8  \\
8  & Myelocyte             & MY  & 1137  & 7  \\
9  & Platelet              & PLT & 2348  & 14 \\
10 & Pro-myelocyte         & PMY & 592   & 4  \\
11 & Segmented Neutrophil  & SNE & 1646  & 10 \\
\midrule
   & \textbf{Total} &      & \textbf{16891} & \textbf{100} \\
\bottomrule
\end{tabular}
\end{table*}

We introduced a twelfth class, Red Blood Cells (RBCs), as an independent category. This addition is essential for comprehensive object detection, as RBCs are the most prevalent cell type, constituting over 90\% of all cells. The complete distribution of the 298,850 cell masks across all twelve classes is shown in Table \ref{tab:pbcmasks_fullwords}.

The original set of 16,891 peripheral blood cell (PBC) images (detailed in Table \ref{tab:cell_distribution 11 categories}) contains these 298,850 individual cell instances.

\begin{table}[htbp]
\centering
\caption{Collected Masks from the PBC dataset}
\label{tab:pbcmasks_fullwords}
\begin{tabular}{lrrr}
\toprule
\textbf{\#} & \textbf{Class} & \textbf{Masks} & \textbf{Percent (\%)} \\
\midrule
1 & Basophil & 1,192 & 0.40 \\
2 & Banded Neutrophil & 1,694 & 0.57 \\
3 & Eosinophil & 3,143 & 1.05 \\
4 & Erythroblast & 1,763 & 0.59 \\
5 & Lymphocyte & 1,246 & 0.42 \\
6 & Metamyelocyte & 1,057 & 0.35 \\
7 & Monocyte & 1,450 & 0.49 \\
8 & Myelocyte & 1,151 & 0.39 \\
9 & Platelet & 14,652 & 4.90 \\
10 & Promyelocyte & 638 & 0.21 \\
11 & Segmented Neutrophil & 1,754 & 0.59 \\
12 & Red Blood Cell & 269,110 & 90.05 \\
\midrule
\textbf{Total} & & \textbf{298,850} & \textbf{100} \\
\bottomrule
\end{tabular}
\end{table}

\subsubsection{Relabeling for Object Detection Using MakeSense.ai }
To enable precise per-cell localization and classification—a necessity for clinical diagnostic applications—the dataset was transformed from an image-level classification format to an object detection framework. This crucial transition was executed using MakeSense~\cite{musleh2023image}, an open-source tool for image annotation. The process involved meticulously drawing tight bounding boxes around every individual cell instance across all 12 refined classes and assigning the corresponding class label. This labor-intensive manual annotation is paramount, as it generates the high-fidelity ground truth data required to train models to accurately classify and localize cells within complex microscopic fields.

\begin{figure}[htbp]
\centering
\includegraphics[width=1\linewidth,height=0.5\textheight,keepaspectratio]{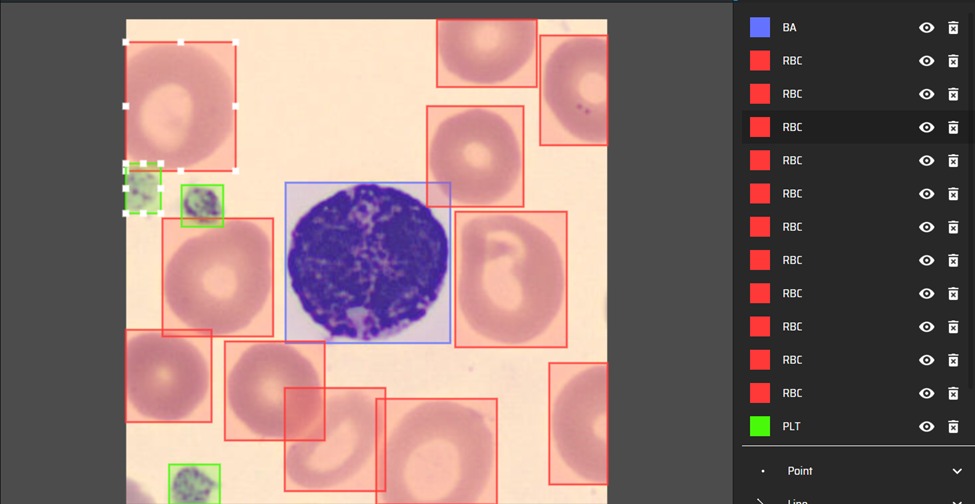}
\caption{Example of a peripheral blood smear image annotated with bounding boxes for red blood cells (RBCs, red), white blood cells (WBCs, blue), basophils (BA, dark blue), and platelets (PLT, green) for object detection analysis.\label{Figure 4}}
\end{figure}

The annotation workflow was standardized using MakeSense.ai to ensure consistency. To guarantee label quality and minimize human error, a rigorous cross-verification protocol was implemented, with multiple annotators reviewing a subset of images to achieve high inter-annotator agreement. The resulting annotations provide a rich dataset where each cell is precisely localized (e.g., 'lymphocyte at [x1, y1, x2, y2]'), a significant enhancement over the original global image labels (e.g., 'image contains lymphocytes'). All annotations were exported in both Pascal VOC (XML) and YOLO (TXT) formats to ensure compatibility with standard object detection architectures. A representative example of the annotated dataset, showcasing color-coded bounding boxes for various cell types, is presented in Figure \ref{Figure 4}.
\subsection{Annotation Transformation for Interoperability}
To maximize the utility, interoperability, and long-term accessibility of the curated dataset, the annotated data was programmatically converted into three standard machine learning formats. This transformation ensures compatibility with a wide range of current and future object detection frameworks and facilitates different stages of the research workflow.

The PASCAL VOC (Visual Object Classes) XML format~\cite{everingham2010pascal} was generated to provide a structured, human-readable record of annotations, ideal for archival and use with numerous deep learning libraries. For model training, the annotations were converted to the native TXT format~\cite{redmon2016you} required by the YOLO (You Only Look Once) family of models, which provides a compact representation using normalized bounding box coordinates. Finally, a CSV (Comma-Separated Values) format~\cite{singh2020tensorflow} was created to enable efficient data inspection, filtering, and analysis using tabular data tools like Pandas, supporting streamlined data management and collaborative research.
All conversions were executed using automated scripts to guarantee accuracy and eliminate manual error, thereby preserving the integrity of the meticulously annotated ground truth data.
\subsection{Dataset Splitting Strategy}
Effective dataset partitioning is critical for developing generalizable models and mitigating overfitting. To rigorously evaluate the robustness and performance of the YOLOv11 variants under different data availability scenarios, two distinct stratified splitting strategies were employed. Stratified sampling was applied in both cases to preserve the original class distribution across all subsets, ensuring a representative and unbiased evaluation.

The first strategy utilized a conventional 7:2:1 split, allocating 70\% of the images for training, 20\% for validation, and 10\% for a hold-out test set. This configuration provides a substantial training corpus for feature learning, a dedicated validation set for hyperparameter tuning and early stopping, and a completely independent test set for the final, unbiased assessment of model performance on unseen data.

A second strategy employed an 8:1:1 split to investigate the impact of increased training data volume on model performance. This approach prioritizes maximum feature exposure during training by allocating a larger portion of the data (80\%), while maintaining a sufficient validation set (10\%) for monitoring generalization and a consistent, independent test set (10\%) for fair comparative analysis between the two strategies. The unchanged test set size ensures that performance comparisons are equitable and not influenced by variations in test set difficulty or composition.
\subsection{Model Selection, Architecture, and Training Configuration}
The YOLOv11 (You Only Look Once version 11) object detection framework~\cite{hidayatullah2025yolo} was selected as the core architecture for this study due to its state-of-the-art performance in real-time object detection and its optimal balance between inference speed and accuracy, which is particularly suitable for medical image analysis applications requiring both precision and efficiency. To comprehensively evaluate the performance-efficiency trade-offs across different model complexities, all five primary variants of the architecture—Nano (n), Small (s), Medium (m), Large (l), and XLarge (x)~\cite{alquraishi2024comprehensive}—were implemented and compared.

\begin{figure*}[htbp]
\centering
\includegraphics[width=.95\linewidth,height=0.5\textheight,keepaspectratio]{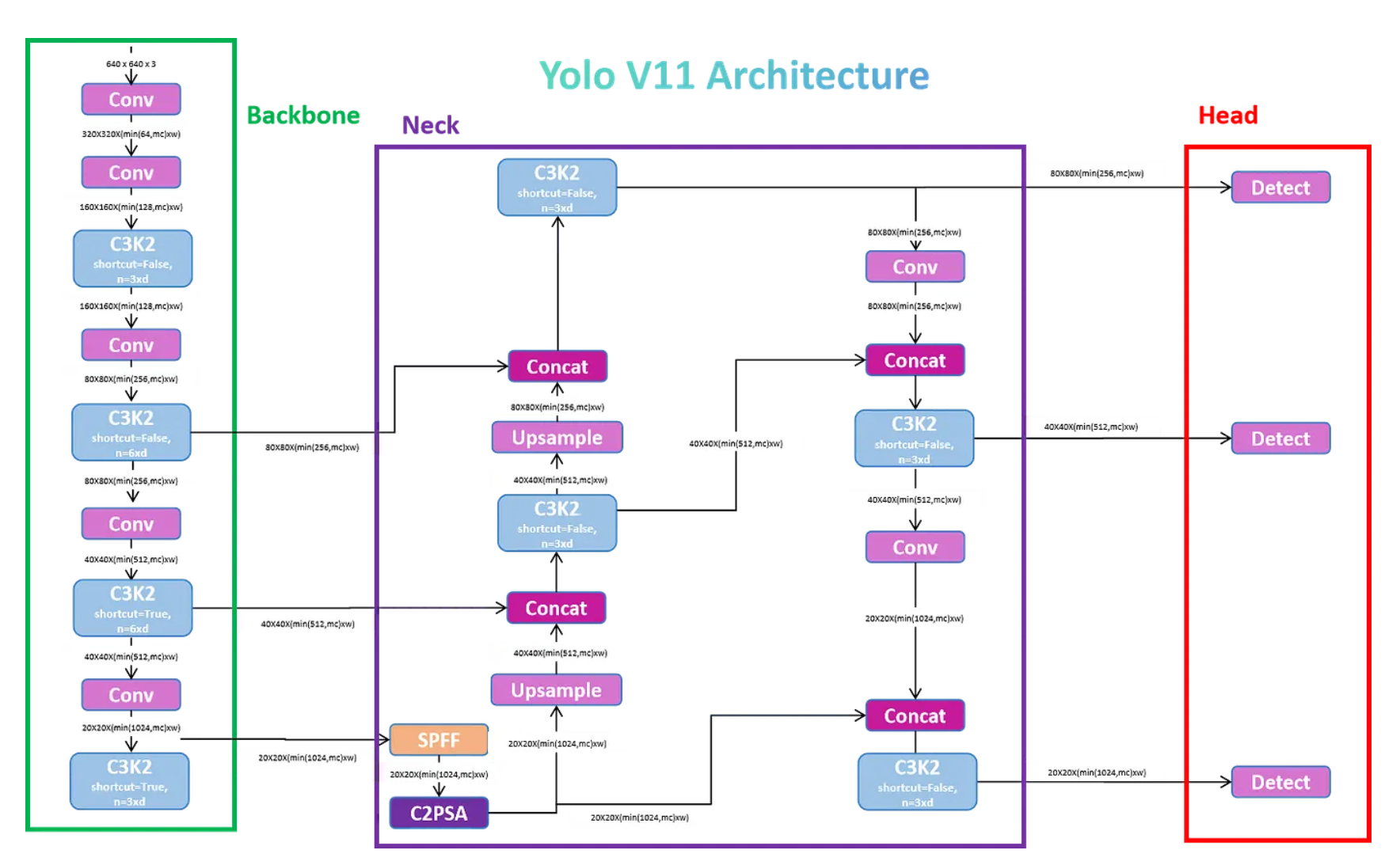}
\caption{YOLOv11 architecture, showing convolutional backbone, spatial pyramid pooling, and anchor-free detection heads}
\label{Figure 5}
\end{figure*}

Each variant was initialized with pre-trained weights from the MS COCO (Common Objects in Context) dataset, leveraging transfer learning to accelerate convergence and enhance feature extraction capabilities on the specialized peripheral blood cell (PBC) dataset. The architectural foundation common to all variants consists of three main components, as illustrated in Figure \ref{Figure 5}: a CSPDarknet backbone for hierarchical feature extraction, a feature pyramid network (FPN) and path aggregation network (PAN) neck for multi-scale feature fusion, and an anchor-free detection head that generates bounding box coordinates, objectness scores, and class probability predictions.

The model variants represent a progressive scaling of this fundamental architecture, with increasing depth, width, and feature capacity from Nano to XLarge. The YOLOv11-Nano variant, designed for edge deployment, prioritizes computational efficiency with minimal parameters, potentially at the cost of discriminating morphologically similar cell types. The Small variant offers a balanced performance-efficiency profile suitable for most clinical applications. The Medium variant provides enhanced representational capacity for capturing nuanced cellular features, while the Large and XLarge variants target maximum accuracy in resource-abundant environments, aiming to address the most challenging cases of cellular differentiation.

All models were trained using a consistent, optimized set of hyperparameters to ensure fair comparative analysis. The training regimen employed stochastic gradient descent with momentum (0.937) and weight decay (0.0005) regularization, with an initial learning rate of 0.01 following a cosine annealing schedule. Extensive data augmentation strategies were implemented, including Mosaic augmentation (probability=1.0), random horizontal flipping (probability=0.5), color space adjustments in HSV domains (hue: ±0.015, saturation: ±0.7, value: ±0.4), and RandAugment automatic augmentation. Models were trained for 100 epochs with a batch size of 16 at 640×640 resolution, using automatic mixed precision to accelerate training while maintaining numerical stability. Early stopping with a patience of 15 epochs was implemented to prevent overfitting, with comprehensive validation performed after each epoch.

A detailed summary of the primary hyperparameters is provided in Table \ref{tab:yolov11-hyperparameters}.

\begin{table*}[htbp]
\centering
\caption{Hyperparameter Configuration for YOLOv11 Training}
\label{tab:yolov11-hyperparameters}
\resizebox{0.8\textwidth}{!}{%
\begin{tabular}{llll}
\toprule
\textbf{Category} & \textbf{Parameter} & \textbf{Value} & \textbf{Function} \\
\midrule
\textbf{Training Strategy} & epochs & 100 & Maximum training iterations \\
& batch & 16 & Batch size for gradient computation \\
& imgsz & 640 & Input image resolution \\
& patience & 15 & Early stopping tolerance \\
\midrule
\textbf{Optimization} & optimizer & SGD & Stochastic Gradient Descent with momentum \\
& lr0 & 0.01 & Initial learning rate \\
& lrf & 0.0001 & Final learning rate (cosine annealing) \\
& momentum & 0.937 & SGD momentum parameter \\
& weight\_decay & 0.0005 & L2 regularization coefficient \\
\midrule
\textbf{Data Augmentation} & mosaic & 1.0 & Mosaic augmentation probability \\
& fliplr & 0.5 & Horizontal flip probability \\
& hsv\_h & 0.015 & Hue augmentation range \\
& hsv\_s & 0.7 & Saturation augmentation range \\
& hsv\_v & 0.4 & Value augmentation range \\
& auto\_augment & randaugment & Automatic augmentation policy \\
\midrule
\textbf{Hardware} & amp & true & Automatic Mixed Precision enabled \\
& workers & 8 & Data loading threads \\
\bottomrule
\end{tabular}%
}
\end{table*}

\subsection{Detection-Oriented Evaluation Metrics}
The performance of all YOLOv11 variants was rigorously evaluated using standard object detection metrics to facilitate a comprehensive comparison across model sizes and data splitting strategies. This multi-faceted assessment ensures a balanced interpretation of each model's capabilities beyond a single performance indicator.
The primary metric was Mean Average Precision (mAP)~\cite{chen2021end}, which provides a holistic measure of both precision and recall across various Intersection over Union (IoU) thresholds. We report two key mAP variants:

•	mAP@0.5: The average precision at an IoU threshold of 0.5, serving as the primary indicator of overall detection accuracy under a common localization criterion.

•	mAP@0.5:0.95: The mean average precision computed over IoU thresholds from 0.5 to 0.95 (in increments of 0.05), providing a stringent assessment of bounding box localization accuracy.

To deconstruct the constituents of mAP and provide interpretability for clinical deployment, we also report precision, recall, and F1-score~\cite{chicco2024improving}. Precision, the ratio of true positives to all positive detections, quantifies the model's reliability and its propensity for false alarms. Recall, the ratio of true positives to all actual positives in the dataset,measures the model's sensitivity and ability to identify all relevant cell instances. The F1-score, as the harmonic mean of precision and recall, offers a single balanced metric that is particularly informative for evaluating performance on rare cell classes amidst the inherent dataset imbalance.

This suite of metrics collectively enables a nuanced analysis of the trade-offs between detection accuracy, localization precision, and classification reliability. The insights derived are critical for guiding model selection based on practical clinical requirements, where the balance between high recall (minimizing missed cells) and high precision (minimizing misdiagnoses) is paramount.
A classification report of confusion matrix~\cite{dalianis2018clinical} evaluates prediction quality using precision, recall, and F1-score per class, along with macro and weighted average accuracies. Accuracy~\cite{dalianis2018clinical}, calculated as a percentage of correct predictions, is determined by Equation  \ref{eq:accuracy}:
\begin{equation}
\text{Accuracy} = \frac{TP + TN}{TP + TN + FP + FN} \label{eq:accuracy}
\end{equation}

Precision~\cite{dalianis2018clinical} measures the quality of a positive prediction made by the model and the Equation  \ref{eq:precision} demonstrates its computational process:
\begin{equation}
\text{Precision} = \frac{TP}{TP + FP} \label{eq:precision}
\end{equation}

Recall~\cite{dalianis2018clinical} measures how many of the true positives (TPs) were recalled (found) and calculated using the Equation \ref{eq:recall}:
\begin{equation}
\text{Recall} = \frac{TP}{TP + FN} \label{eq:recall}
\end{equation}

F1-Score~\cite{dalianis2018clinical} is the harmonic mean of precision and recall and can be calculated using the Equation  \ref{eq:f1}:
\begin{equation}
F1 = 2 \times \frac{\text{Precision} \times \text{Recall}}{\text{Precision} + \text{Recall}} \label{eq:f1}
\end{equation}

\section{Results \& Discussion}
\label{sec:results}

This section presents a comprehensive evaluation of YOLOv11 model variants for automated peripheral blood cell detection and classification, analyzing performance metrics across architectures and data splitting strategies to identify optimal configurations for clinical hematology applications.

\subsection{YOLOv11-Nano Model Performance}
\begin{figure}[htbp]
\centering
\includegraphics[width=\linewidth,height=\textheight,keepaspectratio]{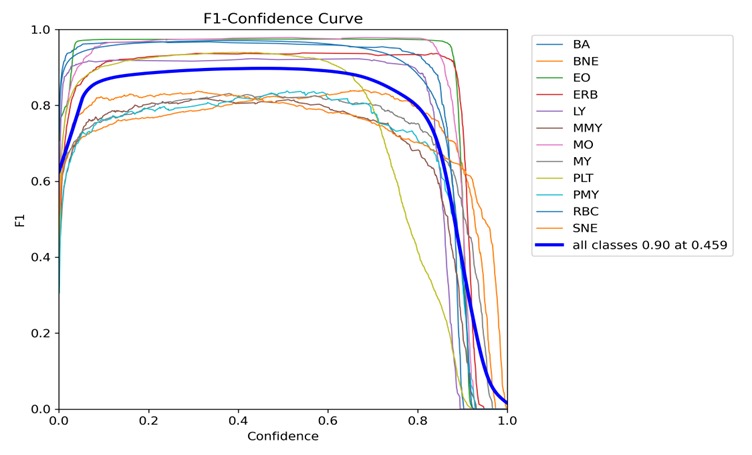}
\caption{F1-Confidence Curve for YOLO Nano model across all 12 PBC classes (7:2:1 Split).\label{Figure 6}}
\end{figure}

As the most lightweight architecture, the YOLOv11-Nano model established a performance baseline. On the 7:2:1split, the model achieved a final mAP@0.5 of 0.878. A clear dichotomy in performance was observed. Morphologically distinct cell types, such as Eosinophils (EO), Monocytes (MO), and Basophils (BA), were detected with high reliability, as evidenced by F1-scores consistently above 0.7, illustrated in the F1-confidence curve Figure \ref{Figure 6}. 

\begin{figure}[htbp]
\centering
\includegraphics[width=0.95\linewidth,height=0.95\textheight,keepaspectratio]{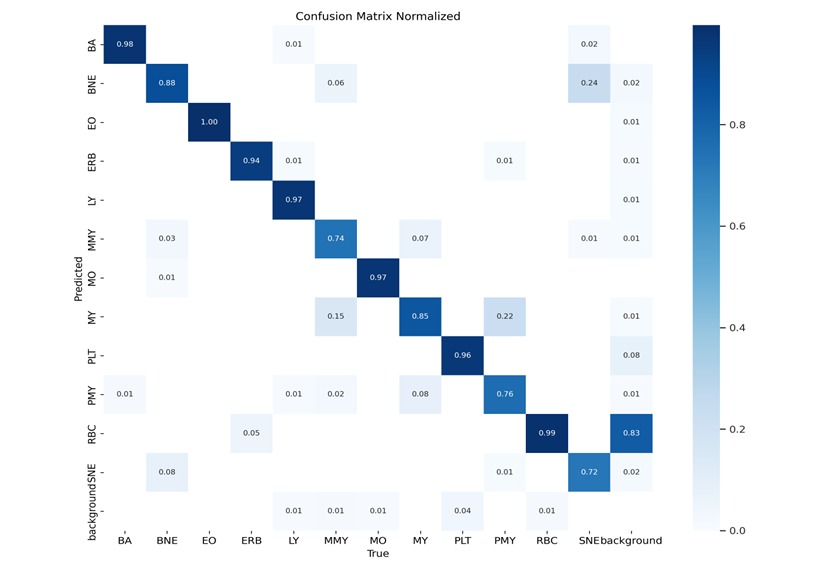}
\caption{Normalized Confusion Matrix for YOLO Nano model (7:2:1 Split) .\label{Figure 7}}
\end{figure}

Conversely, the model exhibited notable difficulty in fine-grained differentiation, particularly among morphologically similar cell types within the granulocytic lineage. Immature granulocytes—promyelocytes (PMY), myeloblasts (MY), and metamyelocytes (MMY)—as well as neutrophil subclasses—banded (BNE) and segmented (SNE)—were frequently misclassified due to their overlapping visual features. This challenge is clearly reflected in the normalized confusion matrix Figure \ref{Figure 7}, which highlights substantial misclassification rates, including PMY misidentified as MY (15.7\%) and MMY as MY (18.3\%). Additionally, platelets (PLT) demonstrated high recall at lower confidence thresholds, yet precision declined sharply at higher thresholds, suggesting persistent confusion with small cellular artifacts.Although the dataset is explicitly defined with twelve biologically relevant blood cell categories, the normalized confusion matrix includes an additional category—background—as part of the evaluation process. This arises from the nature of object detection frameworks such as YOLO, which are trained not only to distinguish among annotated object classes but also to differentiate objects from non-object regions within an image. Any region that does not correspond to a labeled blood cell is implicitly treated as background, allowing the model to suppress false positives and enhance localization accuracy. As a result, the background class is incorporated during evaluation to reflect the model’s ability to correctly identify true cellular instances while avoiding misclassification of irrelevant image regions. This inclusion provides a more comprehensive and diagnostically meaningful assessment of detection performance.

Utilizing the larger training proportion of the 8:1:1 split provided a noticeable performance boost for the Nano model, with its mAP@0.5 rising to 0.892. Improvements were most pronounced in the F1-confidence curve, where the overall macro F1-score peaked at 0.90 (figure 8). Classes such as Erythroblasts (ERB) and Red Blood Cells (RBC) showed more stable performance across confidence thresholds. However, the model's fundamental limitations in distinguishing morphologically similar cell types persisted, as confirmed by the persistent patterns of misclassification visible in the corresponding confusion matrix.
\begin{figure}[htbp]
\centering
\includegraphics[width=\linewidth,height=\textheight,keepaspectratio]{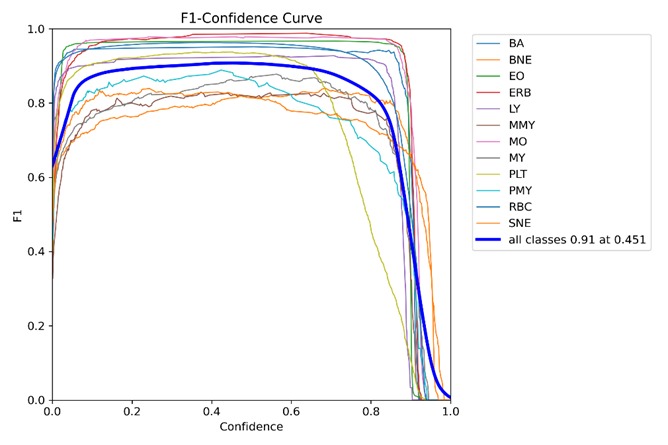}
\caption{F1-score curve for YOLO Nano model showing performance (8:1:1 Split).\label{Figure 8}}
\end{figure} 
\subsection{YOLOv11-Small Model Performance}

\begin{figure}[htbp]
\centering
\includegraphics[width=\linewidth,height=\textheight,keepaspectratio]{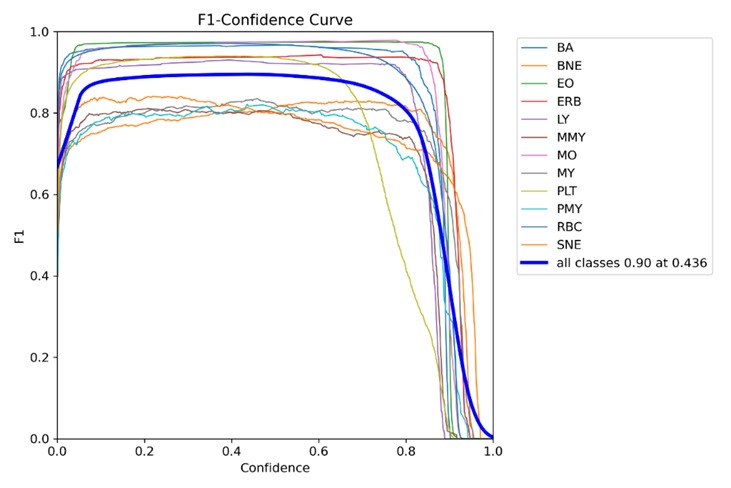}
\caption{F1-score curve for YOLO Nano model showing performance (7:2:1 Split).\label{Figure 9}}
\end{figure}

The YOLOv11-Small model, offering increased capacity, consistently outperformed the Nano variant, achieving a superior balance between computational efficiency and detection accuracy. On the 7:2:1 split, the Small model attained a higher mAP@0.5 of 0.904. The F1-confidence curves Figure \ref{Figure 9} showed modest but consistent improvements across all classes, with the most notable gains observed for the challenging immature granulocytes and neutrophils. 

\begin{figure}[htbp]
\centering
\includegraphics[width=\linewidth,height=\textheight,keepaspectratio]{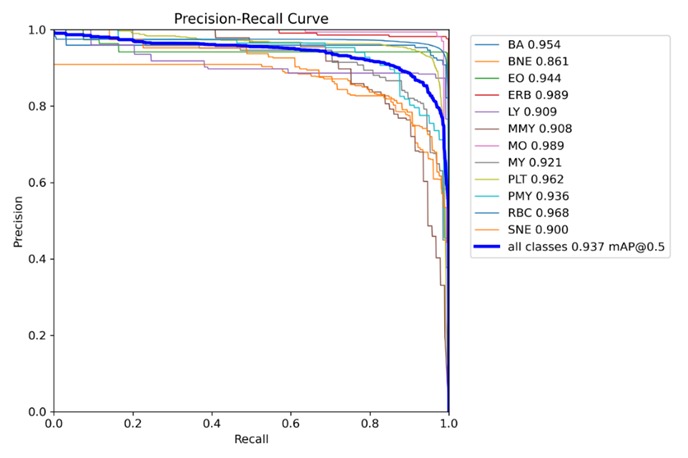}
\caption{Precision-Recall curve for YOLO Small model showing the trade-off between precision and recall across different confidence thresholds (8:1:1 Split).\label{Figure 10}}
\end{figure}

This enhanced discriminatory power is further demonstrated by the normalized confusion matrix, which showed reduced misclassification rates between BNE and SNE, as well as less confusion of PLT with artifacts.
When trained on the 8:1:1 split, the Small model's performance peaked at a mAP@0.5 of 0.915. It achieved a superior precision-recall balance compared to the Nano model, reflected in an overall mAP@0.5 of  0.915. The Precision-Recall curve Figure  \ref{Figure 10} indicates strong performance across most classes, with exceptionally high Average Precision (AP) values for MO (0.989) and ERB (0.989). 

\begin{figure}[htbp]
\centering
\includegraphics[width=\linewidth,height=\textheight,keepaspectratio]{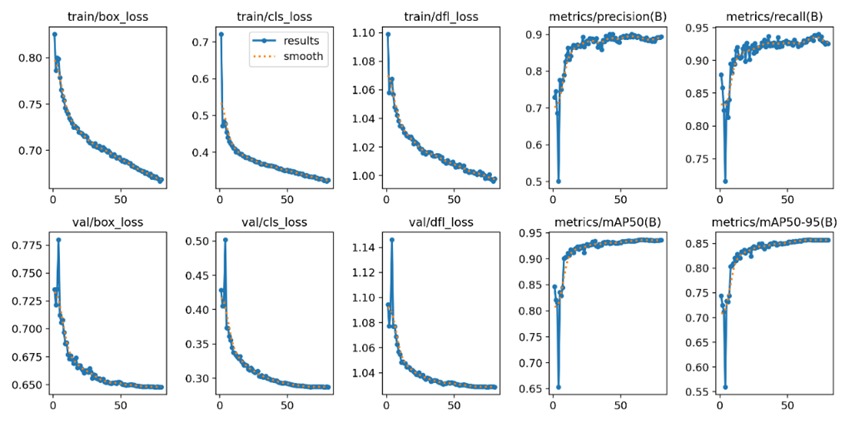}
\caption{Training Results for YOLO Small model (8:1:1 Split).\label{Figure 11}}
\end{figure}

The training process was also more efficient, with convergence occurring at 80 epochs compared to the 92 required for the Nano model, highlighting its optimized learning dynamics, as shown in the training results plot Figure \ref{Figure 11}.

\subsection{YOLOv11-Medium Model Performance}
\begin{figure}[htbp]
\centering
\includegraphics[width=\linewidth,height=\textheight,keepaspectratio]{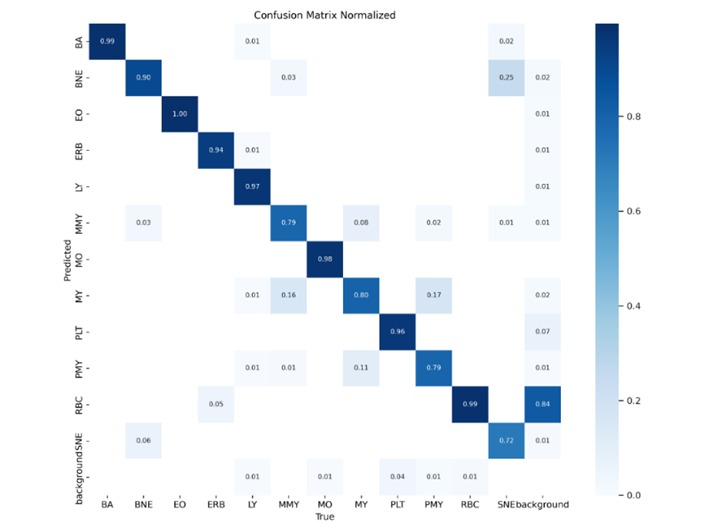}
\caption{Normalized Confusion Matrix for YOLO Medium model (7:2:1 Split).
\label{Figure 12}}
\end{figure}

The YOLOv11-Medium model represented a significant turning point in the performance hierarchy, delivering substantial gains that often surpassed the incremental improvements seen in subsequent larger variants. On the 7:2:1 split, the Medium model delivered a dramatic improvement, achieving an mAP@0.5 of 0.943. This enhancement was most evident in the classification of the most challenging cells. The normalized confusion matrix Figure \ref{Figure 12} reveals a stark reduction in misclassification among immature granulocytes and neutrophils; for instance, the PMY→MY error rate was nearly halved compared to the Nano model. The F1-scores for these classes saw significant increases, indicating the model's augmented capacity is critical for capturing subtle morphological differences.

\begin{figure}[htbp]
\centering
\includegraphics[width=\linewidth,height=\textheight,keepaspectratio]{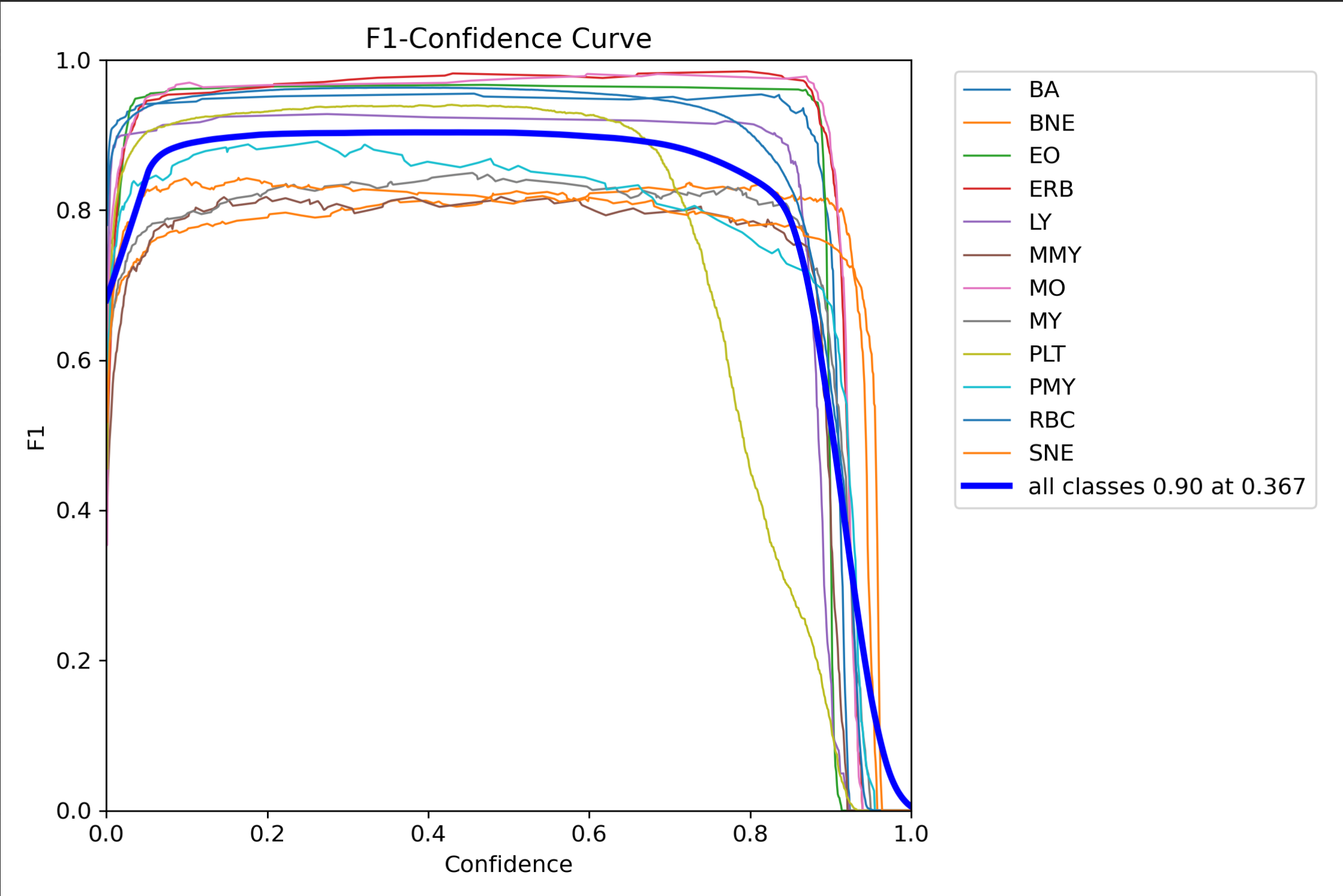}
\caption{F1 confidence medium model (8:1:1  Split).\label{Figure 13}}
\end{figure}

The model maintained its strong performance on the 8:1:1  split, achieving an mAP@0.5 of 0.934. The F1-confidence curve (Figure \ref{Figure 13}) showed further gains for immature granulocytes (e.g., PMY peak ~0.83 vs ~0.80 on Small). The model excelled particularly in recall (0.935), making it highly suitable for screening applications where minimizing false negatives is critical. 

Analysis of the confusion matrix confirmed these findings, showing improved differentiation for all cell types, with PLT misclassification with artifacts dropping to 15.8\%.

\subsection{YOLOv11-Large and XLarge Models Performance}
\begin{figure}[htbp]
\centering
\includegraphics[width=\linewidth,height=\textheight,keepaspectratio]{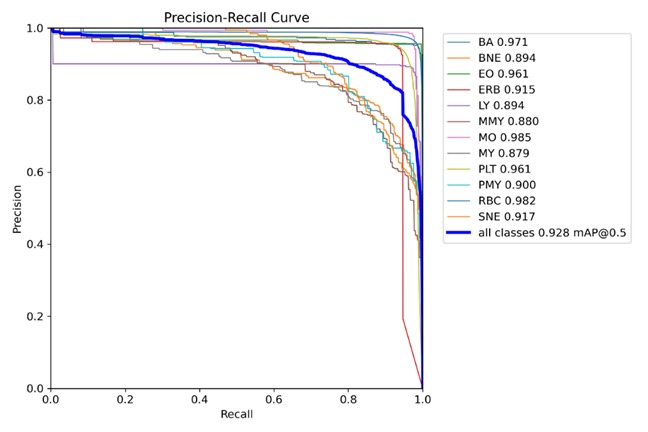}
\caption{Precision-Recall Curve for YOLO Large model (7:2:1 Split).\label{Figure 14}}
\end{figure}

\begin{figure}[htbp]
\centering
\includegraphics[width=\linewidth,height=\textheight,keepaspectratio]{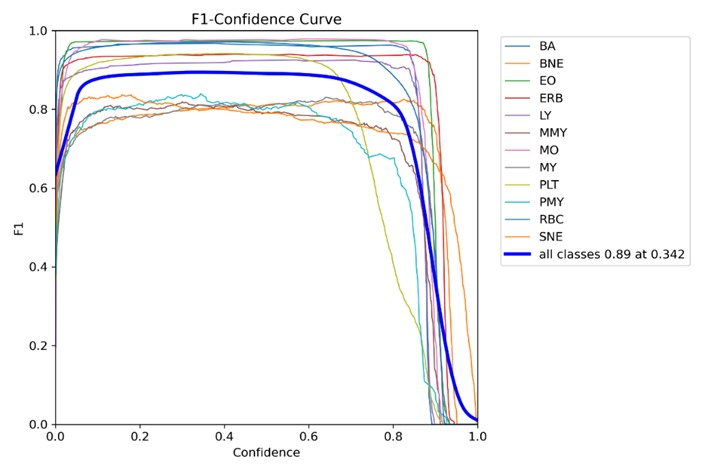}
\caption{F1-Confidence Curve for YOLO XLarge model across all 12 PBC classes (7:2:1 Split).\label{Figure 15}}
\end{figure}

The YOLOv11-Large and XLarge models demonstrated the law of diminishing returns, whereby performance gains became marginal and were not commensurate with the substantial increase in computational cost and complexity.
On the 7:2:1 split, the Large model's mAP@0.5 of 0.945 and the XLarge's value of 0.943 were only marginally higher than the Medium model's 0.943. Analysis of the precision-recall curves for the Large model Figure \ref{Figure 14} and the F1-confidence curves for the XLarge model Figure \ref{Figure 15} revealed nearly identical performance between the Large, XLarge, and Medium models. The confusion matrices for these large variants were also virtually indistinguishable from the Medium model's, confirming that the additional millions of parameters did not translate to meaningfully better feature extraction or classification for this specific task.

\begin{figure}[htbp]
\centering
\includegraphics[width=\linewidth,height=\textheight,keepaspectratio]{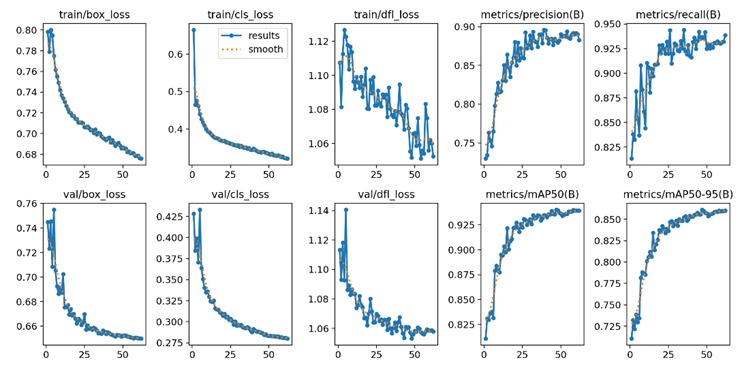}
\caption{Training Results for YOLO XLarge model (8:1:1 Split).}\label{Figure 16}
\end{figure}

This trend continued on the 8:1:1 split. The Large model achieved an mAP@0.5 of 0.947, while the XLarge model reached 0.948—a negligible 0.1\% improvement incurred at a great computational expense. The training results for the XLarge model Figure \ref{Figure 16} showed convergence dynamics and final metric values that were highly similar to those of the Large and Medium models, reinforcing the conclusion that model performance had saturated at the Medium architecture level.

\subsection{Comparative Analysis Across Architectures and Splits}

A holistic comparison of all models and data splits, summarized in Table \ref{tab:yolov11_results}, yields several critical insights. The 8:1:1 data split (80\% training data) consistently outperformed the 7:2:1 split across all models and evaluation metrics. Smaller models (Nano, Small) benefited more substantially from the larger training dataset, while larger models (Medium and above) showed smaller gains, as their inherent capacity was already substantial. Furthermore, performance gains plateaued at the Medium model variant. The minimal increase in mAP from Medium to XLarge was accompanied by a drastic increase in computational cost, both in training time and inference speed. Consequently, the YOLOv11-Medium model trained on the 8:1:1 data split emerged as the optimal choice, achieving 98.6\% of the Large model's performance while requiring significantly fewer computational resources, thereby offering the best practical balance for clinical deployment.

It should be noted that the YOLOv11-XLarge model was trained and evaluated on an NVIDIA A100 GPU, while the Nano, Small, Medium, and Large variants were executed on an NVIDIA T4 GPU. This hardware discrepancy accounts for the unexpected observation where the XLarge model exhibits comparatively shorter training and inference times despite its larger parameter count and higher computational complexity. Consequently, direct runtime comparisons across all variants should be interpreted with caution, as the results are influenced not only by model architecture but also by differences in computational hardware.

\begin{table*}[htbp]
\centering
\caption{Performance comparison of YOLOv11 variants on PBC dataset (splits: 811 and 721).}
\label{tab:yolov11_results}
\begin{tabular}{lcccccc}
\toprule
\textbf{Metric} & \textbf{Split} & \textbf{Nano} & \textbf{Small} & \textbf{Medium} & \textbf{Large} & \textbf{XLarge} \\
\midrule
PARAMS (M) & - & 2.6 & 9.4 & 20.1 & 25.3 & 86.9 \\
\midrule
\multirow{2}{*}{mAP50-95} 
    &8:1:1 & 0.892 & 0.915 & 0.934 & 0.947 & 0.948 \\
    &  7:2:1 & 0.845 & 0.853 & 0.851 & 0.850 & 0.851 \\
\midrule
\multirow{2}{*}{Precision} 
    & 8:1:1 & 0.887 & 0.908 & 0.926 & 0.941 & 0.943 \\
    &  7:2:1 & 0.892 & 0.887 & 0.878 & 0.882 & 0.877 \\
\midrule
\multirow{2}{*}{Recall} 
    &8:1:1 & 0.895 & 0.921 & 0.939 & 0.952 & 0.953 \\
    & 721 & 0.910 & 0.909 & 0.919 & 0.916 & 0.921 \\
\midrule
\multirow{2}{*}{F1-Score (\%)} 
    & 8:1:1 & 0.891 & 0.914 & 0.932 & 0.946 & 0.948 \\
    &  7:2:1 & 89.75 & 89.24 & 89.82 & 89.98 & 89.84 \\
\midrule
\multirow{2}{*}{Training Time (hrs)} 
    & 8:1:1& 3.2   & 4.5   & 6.8   & 6.0   & 4.0 \\
    & 7:2:1 & 8.039 & 8.794 & 14.066 & 17.393 & 4.707 \\
\midrule
\multirow{2}{*}{Inference Speed (ms/img)} 
    & 8:1:1 & 12.3 & 18.7 & 27.4 & 42.8 & 50.5 \\
    &  7:2:1 & 2.2  & 4.8  & 10.7 & 14.6 & 3.7 \\
\bottomrule
\end{tabular}
\end{table*}

\subsection{Qualitative Results: Ground-Truth versus Prediction}
\label{sec:qual_results}

This section provides a qualitative analysis of the model performances by directly comparing their predictions against the ground-truth annotations for the test set. This visual assessment complements quantitative metrics by highlighting specific strengths, such as accurate classification and localization, and weaknesses, such as misclassifications between morphologically similar cell types.

\begin{figure}[htbp]
    \centering
    \includegraphics[width=\linewidth,keepaspectratio]{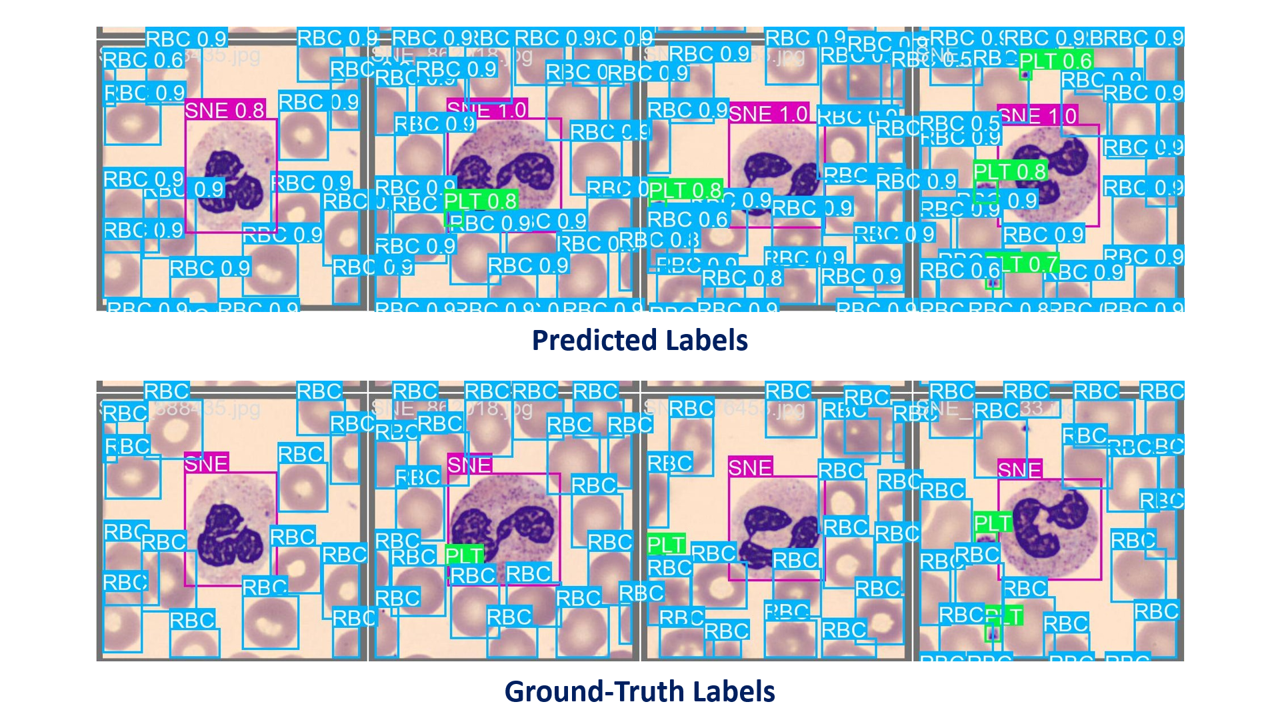}
    \caption{Exemplary detection results from the YOLOv11 Small model (8:1:1 Split). The top row shows the model's predictions with bounding boxes and confidence scores for Segmented Neutrophils (SNE), Red Blood Cells (RBC), and Platelets (PLT). The bottom row provides the corresponding ground-truth annotations.}
    \label{fig:comparison_sne}
\end{figure}

Figure \ref{fig:comparison_sne} demonstrates the robust performance of the YOLOv11 Small model. The visual comparison reveals precise alignment between the predicted bounding boxes and the ground-truth annotations for Segmented Neutrophils (SNE), Red Blood Cells (RBC), and Platelets (PLT). The high confidence scores associated with these correct predictions indicate the model's certainty in detecting these common cell types.

\begin{figure}[htbp]
    \centering
    \includegraphics[width=\linewidth,keepaspectratio]{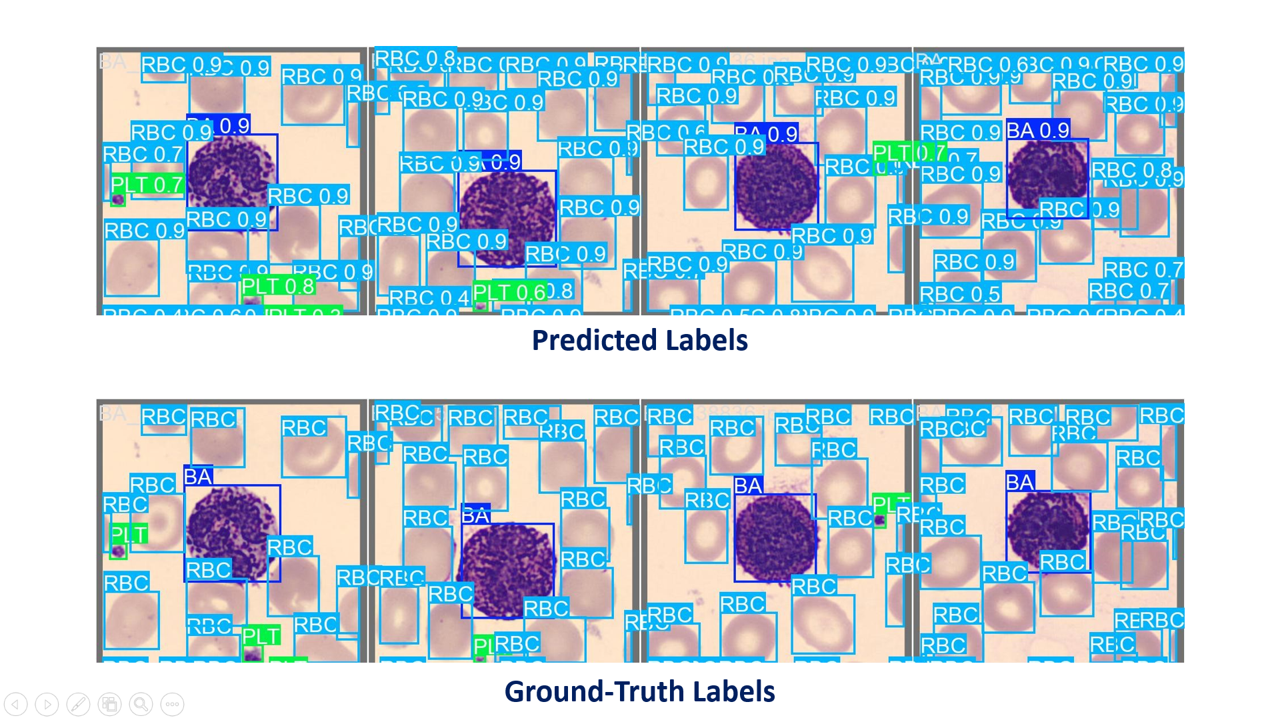}
    \caption{Detection results for the YOLOv11 Nano model (8:1:1 Split) on images containing Basophils (BA). The top row shows the model's output, which correctly identifies BA, RBC, and PLT. The bottom row confirms the accuracy of these predictions.}
    \label{fig:comparison_ba}
\end{figure}

This high standard of performance is also evident in the results from the smaller YOLOv11 Nano model. As shown in Figure \ref{fig:comparison_ba}, the Nano variant successfully identifies and localizes the less frequent Basophils (BA) in addition to RBCs and Platelets. The consistency between its predictions and the ground-truth labels confirms that this compact model also generalizes effectively on the PBC dataset for these specific classes.

\begin{figure}[htbp]
    \centering
    \includegraphics[width=\linewidth,keepaspectratio]{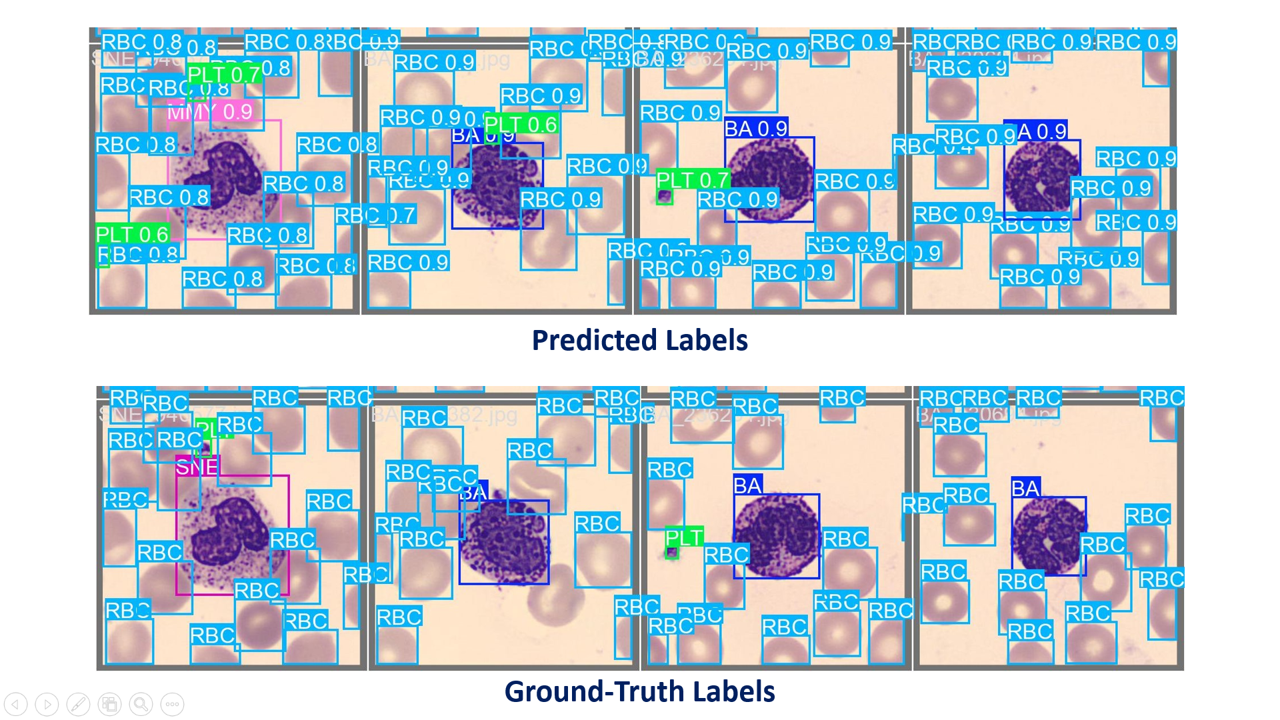}
    \caption{A case study of a misclassification by the YOLOv11 Nano model (8:1:1 Split). The model correctly identifies RBCs and PLTs but misclassifies a Segmented Neutrophil (SNE) as a Metamyelocyte (MMY) with high confidence (0.9).}
    \label{fig:comparison_misclassification}
\end{figure}

However, not all predictions are flawless, and qualitative analysis is key to identifying specific failure modes. Figure \ref{fig:comparison_misclassification} provides a critical example where the YOLOv11 Nano model, while correctly identifying most cells, misclassifies a Segmented Neutrophil (SNE) as a Metamyelocyte (MMY) with a high confidence score of 0.9. This error is highly informative, as it pinpoints a confusion between two developmentally related white blood cells that can appear morphologically similar—a common and challenging nuance in cytological image analysis. This insight is valuable for guiding future work, suggesting a need for targeted data augmentation or a more refined network architecture to improve discriminative ability between such similar classes.

In summary, the qualitative results demonstrate that both YOLOv11 variants perform robustly in detecting and classifying the majority of blood cells, showing high precision in localizing objects and assigning correct labels to distinct classes. The primary area for improvement, as identified by the specific misclassification, lies in enhancing the model's ability to discriminate between specific, similar types of white blood cells to reduce confident false positives.

\subsection{Discussion}
This study presents a comprehensive benchmark of the YOLOv11 architecture family for automated peripheral blood cell detection, demonstrating its strong potential to augment hematological diagnostics. Our central finding identifies YOLOv11-Medium as the Pareto-optimal model, achieving an exceptional balance between high accuracy (mAP@0.5 of 0.934) and computational efficiency, making it the most suitable variant for real-world clinical deployment. The high performance across all variants, particularly the elevated mAP metrics, underscores the capability of deep learning-based object detection to mitigate the labor-intensive and subjective limitations inherent in manual peripheral blood smear examination.

A critical insight from this work is the significant influence of dataset partitioning strategy on model performance. The consistent superiority of the  8:1:1 split over the conventional  7:2:1 split across all models indicates that for complex, imbalanced cytological datasets, maximizing the volume of training data is a more decisive factor for robustness than maintaining a larger validation set for hyperparameter tuning. This finding provides a valuable practical guideline for developing medical imaging AI systems where annotated data is scarce but precious.

Despite these promising results, several limitations must be acknowledged to contextualize our findings. Firstly, the models were trained and validated on images acquired from a single system (CellaVision DM96) under a specific staining protocol, which may impair generalizability to images from other hematology analyzers. Secondly, the absence of external validation on a multi-center dataset limits our ability to assess robustness across diverse patient populations and laboratory conditions. Finally, persistent challenges, such as severe class imbalance and the inherent morphological similarity between certain cell lineages (e.g., immature granulocytes), remain a source of residual misclassification, suggesting that architectural advances alone may be insufficient to fully resolve these ambiguities.

Future research should prioritize multi-institutional collaboration to build large, diverse, and multi-source datasets for external validation, which is the gold standard for demonstrating clinical utility. To address the lingering challenge of inter-class heterogeneity, investigations into advanced domain adaptation techniques, vision transformer-based architectures, and explainable AI (XAI) methods are warranted. These approaches could enhance model generalization and provide clinicians with interpretable diagnostic insights. Ultimately, the translation of these systems into diagnostic workflows requires rigorous clinical usability studies, incorporating a "pathologist-in-the-loop" framework to evaluate the practical synergy between AI-assisted screening and expert morphological interpretation.

In conclusion, this work establishes a robust, evidence-based framework for selecting object detection models in medical imaging, challenging the notion that larger models are invariably superior. By providing a detailed analysis of the accuracy-efficiency trade-offs across the YOLOv11 spectrum, we offer a clear roadmap for integrating scalable AI solutions into clinical hematology. The adoption of such systems holds the potential to standardize peripheral blood smear analysis, enhance diagnostic reproducibility, and streamline laboratory workflows globally, ultimately paving the way for more accessible and consistent hematological diagnostics.

\section{Conclusion}
\label{sec:conclusion}

This study conducted a rigorous evaluation of the YOLOv11 architecture for automated peripheral blood cell detection, systematically comparing five model variants under two data partitioning strategies. Our findings demonstrate that while model capacity is crucial for capturing complex cellular morphology, diminishing returns occur beyond the Medium variant, where larger models yield negligible performance gains despite significantly increased computational demands. Also, An annotated dataset combining detection and classification labels for cell type has been created and released on GitHub to aid future studies.

The YOLOv11-Nano variant provided competent baseline performance but showed limitations in distinguishing morphologically similar cell types. The Small and Medium variants delivered substantial improvements, with the Medium variant emerging as the optimal architecture, achieving near-peak detection accuracy (mAP@0.5 of 0.934) while maintaining practical computational efficiency. Our analysis further revealed that increased training data (8:1:1 split) consistently enhanced performance, with smaller models benefiting disproportionately from additional data.

Based on our comprehensive evaluation, we recommend the YOLOv11-Medium model trained with an  8:1:1 data split as the optimal configuration for clinical deployment. This selection represents the Pareto-optimal balance between accuracy and efficiency, making it suitable for integration into real-time diagnostic workflows.

This work challenges the conventional "larger is better" paradigm in medical AI and provides an evidence-based framework for model selection in hematological image analysis. Our findings represent a significant step toward routine clinical adoption of automated PBC analysis, with potential to enhance diagnostic precision, standardize morphological interpretation, and streamline laboratory workflows globally.

Future research will focus on external validation through multi-center collaborations and development of advanced techniques to address persistent challenges in fine-grained cell differentiation, particularly for rare cell types and immature morphological variants. In addition, different object-identification architectures will be evaluated using our generated dataset.


\end{document}